\documentclass[runningheads]{llncs}
\usepackage[T1]{fontenc}
\usepackage{graphicx}
\usepackage{amsmath}
\usepackage{amssymb}
\usepackage{booktabs}
\usepackage[table]{xcolor}

\newcommand{\f}[1]{\textcolor{red}{#1}}
\newcommand{\s}[1]{\textcolor{blue}{#1}}

\newcommand{\std}[1]{\tiny {$\pm$#1} }

\usepackage{adjustbox}

\begin{document}
\title{Joint Discriminative and Metric Embedding Learning for Person Re-Identification}
%
\titlerunning{Joint Discriminative and Metric Embedding Learning for Person Re-ID}
%
\author{Sinan I. Sabri\inst{1,2} \and
Zaigham A. Randhawa\inst{1} \and
Gianfranco Doretto\inst{1}}
\authorrunning{S. I. Sabri et al.}
%
\institute{West Virginia University, Morgantown, WV 26506, USA\\
\email{\{sisabri,zar00002,gidoretto\}@mix.wvu.edu} \and 
University of Misan,  Amarah, Maysan, Iraq}

\maketitle              
\begin{abstract}
 Person re-identification is a challenging task because of the high intra-class variance induced by the unrestricted nuisance factors of variations such as pose, illumination, viewpoint, background, and sensor noise. Recent approaches postulate that powerful architectures have the capacity to learn feature representations invariant to nuisance factors, by training them with losses that minimize intra-class variance and maximize inter-class separation, without modeling nuisance factors explicitly. The dominant approaches use either a discriminative loss with margin, like the softmax loss with the additive angular margin, or a metric learning loss, like the triplet loss with batch hard mining of triplets. Since the softmax imposes feature normalization, it limits the gradient flow supervising the feature embedding. We address this by joining the losses and leveraging the triplet loss as a proxy for the missing gradients. We further improve invariance to nuisance factors by adding the discriminative task of predicting attributes. Our extensive evaluation highlights that when only a holistic representation is learned, we consistently outperform the state-of-the-art on the three most challenging datasets. Such representations are easier to deploy in practical systems. Finally, we found that joining the losses removes the requirement for having a margin in the softmax loss while increasing performance.

\keywords{Person Re-Identification  \and Discriminative Learning \and Metric Learning.}
\end{abstract}

\section{Introduction}

Person re-identification~\cite{bak2017one,liao2015person,pedagadi2013local,zhang2016learning} is the task of assigning the same identity to tightly cropped images of people, based solely on their whole body appearance information. The problem is challenging because distinct images of the same person may look very different, since no restrictions are imposed on the nuisance factors of variation, such as pose, illumination, viewpoint, background, and sensor noise, causing a high intra-class variance. 

In order to address that challenge, the research landscape has evolved from developing feature-based models~\cite{farenzena2010person,liu2012person} coupled with metric learning~\cite{Wei-Shi_Zheng2011-zy}, to developing dedicated deep learning architectures~\cite{Zheng2016-bm} trained with classification and verification losses~\cite{Li2017-th}, to developing specialized deep learning schemes~\cite{zheng2019joint,yang2019towards} aiming at extracting more robust feature embeddings by leveraging powerful pretrained backbone architectures like ResNet-50~\cite{He2016cvpr}. In addition to that, some recent works \cite{lin2019improving,zhang2018person,tay2019aanet,chikontwe2018deep,li2020attributes} used person attributes such as gender, upper and lower body clothing colors, carrying handbag and backpack as a powerful complementary information to improve the performance of person re-identification. These attributes have more discriminative information about person images, and are invariant to nuisance factors, and they could help with coping with intra-class variations.

Among the more recent trends, there has been also the idea of learning feature embeddings directly suitable for re-identification, by improving the ability to control how losses deal with intra-class and inter-class variances, while giving much less importance to the explicit modeling of the nuisance factors of variation. These approaches focus on learning a holistic representation of the image of a person. They are simpler to deploy, and incorporate in a retrieval system. There are two main lines of work. The first one has focussed on improving the triplet loss derived from metric learning~\cite{Weinberger2009jmlr,Hermans2017triplet}. The second line of work has focused on improving the softmax loss used for classification, via normalizing weights and representations~\cite{liu17sphereface,wang18cosface,deng2018arcface}. However, we note that restricting the embeddings to live on a hypersphere limits the gradient flow supervising the embedding under training, which could potentially generate a performance gap.

In this work, we improve the learning of a holistic representation in the form of a feature embedding for person re-identification. Inspired by the previous observation, we do so by incorporating the latest findings on the softmax and triplet losses in a revised combination of such losses, which includes also the learning of multiple discriminative tasks, given by the identity classification and the prediction of attributes. The intent is for the triplet loss to help the softmax further decrease intra-class variation, and increase inter-class distance by letting the triplet loss be the proxy for the gradient supervision that the embedding normalization has restricted, and we specify under what conditions this may happen. We also observe that the same strategy used to form the batch of triplets can be used in tandem with the softmax loss to prevent issues due to dataset imbalance, which are common in person re-identification. In addition, we add the discriminative task of learning attributes to further increase robustness against nuisance factors. We perform an extensive evaluation of the proposed combination of losses with and without using person attributes on the latest person re-identification datasets. We found that this approach can achieve competitive performance with the state-of-the-art, and that the combined loss does not require the softmax component to use any margin.


\section{Related Work}

Person re-identification is a challenging task due to the nuisance factors of variation, such as pose, illumination, viewpoint, background clutter, spatial misalignment, etc. There is a large literature in this area~\cite{Ye2022-lz}, and two predominant directions for tackling the problem are based on metric learning~\cite{bak2017one,hirzer2012relaxed,liao2015person,pedagadi2013local,zhang2016learning}, and discriminative feature representation learning~\cite{farenzena2010person,liu2012person,ma2012bicov}.

Recent works use deep learning to learn robust feature representations~\cite{Ye2022-lz}. \cite{xiao2016learning} proposes to jointly learn features from multiple domains and then finetuning with domain guided drop out for the specific domain. \cite{ahmed2015improved} offers a deep convolutional architecture trained on pairs of images capable of learning features and similarity metric simultaneously. \cite{chen2018group} combines the CRF model with DNN to learn more consistent multi-scale similarity metrics for person re-identification. \cite{sun2018beyond} employs partition strategy on convolutional features, and \cite{fan2019spherereid} learns embedding of the person image on a hypersphere manifold using a spherical loss. Differently from \cite{fan2019spherereid}, our proposed model uses a simpler architecture and focusses on combining a margin based softmax loss with a triplet loss to expand feature embedding hypothesis.  \cite{zhao2017deeply} detects body regions that are discriminative for person re-identification, while \cite{li2017learning} learns full body and body parts features through a multi scale context aware network. \cite{hou2019interaction} addresses the limitation of CNNs in representing person images with large variations in body pose and scale by proposing a module to conclude the receptive fields according to the pose and scale of the input person image. \cite{yang2019towards} explores diverse discriminative visual cues without the assistance of pose estimation and human parsing, and \cite{wang2018mancs} proposes a Fully Attention Block (FAB) plugged into a CNN to overcome the misalignment problem and to localize discriminative local features. Generative adversarial networks (GANs) \cite{goodfellow2014generative} have been used in person re-identification. \cite{ge2018fd,qian2018pose,zheng2019joint} aim at decoupling pose information from image features via adversarial learning.

Person attributes have been used in person re-identification leading to improved robustness against variation of viewpoint, illumination and pose. \cite{lin2019improving} manually annotated person attributes for the Market-1501~\cite{zheng2015scalable} dataset and the DukeMTMC-ReID~\cite{zheng2017unlabeled} dataset. It proposes the attribute-person recognition (APR) network. \cite{zhang2018person} transforms attribute recognition from a high level layer to a mid level layer, and \cite{liu2018ca3net} jointly learns appearance and attribute representations via multi-task learning. To learn discriminative person body parts, \cite{tay2019aanet} utilizes person attribute information by integrating attribute features with identity and body part classification. \cite{chikontwe2018deep} proposes a multi task network to learn identity part-level representation and an attribute global representation.  \cite{li2020attributes} uses person attributes to detect attribute body parts or handle body parts misalignment.


\section{Proposed Approach}

For the person re-identification task, given a tightly cropped image sample of a person, $I$, we are seeking to learn a feature embedding $f_{\theta}(I)$, defined by the set of parameters $\theta$, which is as invariant as possible to the nuisance factors of variations. Rather than attempting to model nuisance factors, current deep neural network architectures have shown the promise to cope with their effects, by shifting the focus on designing clever training practices, as well as loss functions. Here we intend to further explore this holistic-based approach and shed light on additional aspects of this line of work.

\subsection{Classification Losses}

A succesful strategy for learning the embedding $f_{\theta}$ is through the use of a classification loss such as the categorical cross-entropy, which entails adding a softmax layer after the embedding. This leads to the loss 
\begin{equation}
  \mathcal{L}_{S}  (\theta, W, b) = - \frac{1}{N}  \sum_{i=1}^{N} \log \frac{e^{ W_{y_i}^{\top} x_i + b_{y_i}}}{\sum_{j=1}^{n} e^{W_{j}^{\top} x_i + b_j}} \; ,
  \label{eq-softmax}
\end{equation}
where $x_i=f_{\theta}(I_i)\in \mathbb{R}^d$ is the embedding of $I_i$, which has identity $y_i$. Moreover, $W = [W_1, \cdots, W_c]\in \mathbb{R}^{d \times c}$ and $b=[b_1, \cdots, b_c]$ are the weights and biases of the softmax layer, while $N$ is the batch size.

Given two images $I_i$ and $I_j$ of the same identity, i.e., $y_i=y_j$, the softmax loss~\eqref{eq-softmax} will strive to make the target logit in position $y_i$ be the highest for both images. While this should encourage $f_{\theta}(I_i)$ and $f_{\theta}(I_j)$ to be close, in general, there is not an explicit effort to impose $f_{\theta}(I_i) = f_{\theta}(I_j)$. This leads to a performance gap, given the large intra-class variability of the person re-identification task due to nuisance factors, which easily cause identity miss-classifications.
\begin{figure*}[t!]
\begin{center}
  \includegraphics[width=\linewidth]{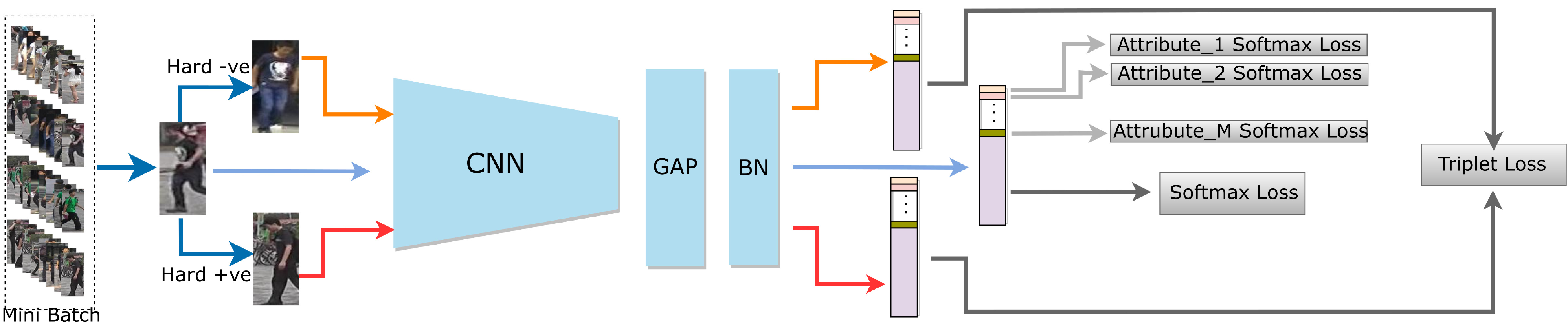}
\end{center}
\caption{\textbf{Architecture.} Simple graphical description of the joint optimization of the multi-task re-identification loss, based on a multi-class classification~\eqref{eq-arcface} (for the identities), a multi-label classification~\eqref{eq-arcface-attr} (for attributes), and a metric learning loss~\eqref{eq-batch-hard}. 
}
\label{fig:architecture}
\end{figure*}

Within the context of face recognition, the issue above has beeen mitigated by taking several steps. First, every logit is produced by comparing the input against $\ell_2$-normalized weights~\cite{liu17sphereface,wang18cosface,wang2017normface}, i.e. $\|W_j
\|=1$. This reduces by one the degrees of freedom by which two different logits could become equal, when activated by images $I_i$ and $I_j$ respectively, each of which depicting the same identity, i.e., $y_i=y_j$. Second, the input of every logit is also $\ell_2$-normalized~\cite{wang18cosface}, i.e. $\|x_i\|=1$, and rescaled to a temperature value $s$. This further reduces the degrees of freedom by which different logits could become equal, by imposing the embeddings to be defined on the hypersphere of radius $s$. Also, this suggests using cosine similarity as the metric for comparison between inputs and weights.

While input and weight normalizations positively contribute towards reducing intra-class variability, it is possible to further reduce the spread of the embeddings of the samples with same identity. This is done by introducing a margin in the cosine similarity, $\cos (\alpha)$, of the target logit. Doing so would further pull the embeddings closer to make up for the loss of similarity induced by the margin.
There are at least three basic ways to add a cosine similarity margin~\cite{liu17sphereface,deng2018arcface,wang18cosface}. In~\cite{deng2018arcface} it is shown that for face recognition the \emph{additive angular} margin $m$ is the most effective, which reduces~\eqref{eq-softmax} to
\begin{equation}
  \mathcal{L}_{AM}  (\theta, W) = - \frac{1}{N}  \sum_{i=1}^{N} \log \frac{e^{ s
      \cos( \alpha_{y_i}+m)}}{e^{ s \cos( \alpha_{y_i}+m)} + \sum_{j\ne y_i} e^{ s \cos \alpha_{j}}} \; ,
  \label{eq-arcface}
\end{equation}
where we have set $b=0$ for simplicity, as in~\cite{liu17sphereface}. In the experiments we explore the effectiveness of~\eqref{eq-arcface} for person re-identification. Its action should be to minimize the intra-class variation, while the denominator attempts to maximize the inter-class discrepancy by distancing the weights on the unit hypersphere.

We further push the training of the embedding to become invariant to the nuisance factors by leveraging the attribute labels. Assuming that a person image $I$ is described by $M$ binary attributes, the original embedding $f_{\theta}(I)$ is now split into the inputs of two heads, $f_{\theta_{id}}$  and $f_{\theta_{a}}$, for predicting identity and attributes, respectively. The first input $f_{\theta_{id}}$ will be trained according to~\eqref{eq-arcface}, which we indicate more specifically as $\mathcal{L}_{AM_{id}}  (\theta_{id}, W_{id})$.

The second input $f_{\theta_{a}}$ for attribute prediction is still normalized and the head weights are normalized as well to minimize intra-class variability and maximize the correct prediction of the attributes. However, since every attribute is binary (i.e., present or not), in order to implement the normalization strategy, and leverage the additive angular margin loss~\cite{deng2018arcface}, we cannot treat this as a multi-label problem where we use the binary cross-entropy loss for every attribute. Instead, we must use a categorical cross-entropy loss for every attribute where the number of categories is 2. Therefore, the corresponding loss for this pool of $M$ classifiers becomes
\begin{equation}
  \mathcal{L}_{AM_{attr}}  (\theta_{a}, W_{a}) = - \frac{1}{N}  \sum_{i=1}^{N} \sum_{k=1}^{M} \log \frac{e^{ s
      \cos( \alpha_{a_{k,i}}+m)}}{e^{ s \cos( \alpha_{a_{k,i}}+m)} + e^{ s \cos \bar\alpha_{a_{k,i}}}} \; ,
  \label{eq-arcface-attr}
\end{equation}
where $\bar \alpha_{a_{k,i}}$ indicates the input to the cosine corresponding to the option where the attribute is not present.
The final multi-task classification loss imposed on the identities and attributes is the sum of $\mathcal{L}_{AM_{id}}  (\theta_{id}, W_{id})$ and~\eqref{eq-arcface-attr}. Also, note that $\theta_{id}$ and $\theta_{a}$ actually share a significant amount of weights, since they only differ for the weights in the last layer, leading up to the two embedding components $f_{\theta_{id}}$  and $f_{\theta_{a}}$. They are indicated in that way to limit notation clutter.

\subsection{Metric Learning Loss}

By learning a metric embedding we directly train a function $f_{\theta}$ that maps images of the same identity as close as possible, effectively minimizing the intra-class variability of the embeddings, while images of different identities are mapped far away, creating a large inter-class discrepancy. In~\cite{Weinberger2009jmlr} they developed a margin based approach for $k$-nearest neighbor classification, which has then inspired the \emph{triplet loss} formulation of FaceNet~\cite{Schroff2015cvpr} as follows
\begin{equation}
  \mathcal{L}_T(\theta) = \!\!\!\!\!\!\!\!\!\!\!\!\!\!\! \sum_{\small (a,p,n) \; \mathrm{s.t.} \; y_a=y_b\ne y_n}   \!\!\!\!\!\!\!\!\!\!\!\! [m+ D(f_{\theta}(I_a),f_{\theta}(I_p)) - D(f_{\theta}(I_a),f_{\theta}(I_n))]_+
  \label{eq-triplet}
\end{equation}
where $D(\cdot,\cdot)$ denotes a suitable distance, and $[\cdot]_+$ is the hinge function, but other surrogates could be used, such as the softplus function $\ln(1+\exp(\cdot))$. The triplet loss~\eqref{eq-triplet} operates by ensuring that the distance between a positive sample $I_p$ and an anchor $I_a$, which have same identities, is smaller than the distance between the ancor and a negative sample $I_n$, which has a different identity, at least by a margin $m$. When the loss is optimized over a large combination of triplets $(I_a,I_p,I_n)$, it pulls embeddings of the same identity, while pushing apart those with different identities.   

The challenges in using the triplet loss are related to the cubic growth in the number of triplets as the dataset size grows, and in forming meaningful triplets. It turns out that the embedding can quickly learn how to correctly map easy triplets. Conversely, focusing on selecting very hard triplets may not be very useful too, because we would teach the embedding how to map outlier cases, while overlooking how to handle well ``average'' cases. This is why it is important to efficiently mine moderate positives and negatives~\cite{Schroff2015cvpr,song16cvpr}.

As described in~\cite{Hermans2017triplet}, it turns out that there is an effective way to address both of the issues above. Triplets can be formed out of selecting $P$ identities, and $K$ samples per identity, with a total of $PK$ samples in a batch. Since we are operating only within a batch, these hard selections will not be outliers, but mostly non-trivial moderate cases. In addition, this approach avoids the overhead induced by mining moderate cases from the full dataset processed by the latest update of $f_{\theta}$. This procedure, named \emph{batch hard}~\cite{Hermans2017triplet}, changes~\eqref{eq-triplet} into
\begin{equation}
  \mathcal{L}_{BH}(\theta) =  \sum_{i=1}^P\sum_{a=1}^K  \;\; [ m  + \max_p D(f_{\theta}(I_a^i),f_{\theta}(I_p^i))  
      - \min_{j,n,j\ne i} D(f_{\theta}(I_a^i),f_{\theta}(I_n^j)) ]_+ \; .
  \label{eq-batch-hard}
\end{equation}

\subsection{Joint Classification and Metric Loss}
\label{subsec:joint-class-metric}

Besides evaluating the additive angular magin softmax loss~\eqref{eq-arcface} and the batch hard triplet loss~\eqref{eq-batch-hard} on the most recent re-identification datasets, we plan to study their contribution into a joint loss
\begin{equation}
  \mathcal{L}_{AMBH} (\theta,W) = \mathcal{L}_{AM}(\theta,W) + \gamma \mathcal{L}_{BH} (\theta) \; .
  \label{eq-joint-loss}
\end{equation}
where $\gamma$ is a hyperparameter balancing the relative strengths of the losses.

There are a couple reasons that motivate the exploration of the join loss~\eqref{eq-joint-loss}. The first one comes from observing that a major drawback of the loss~\eqref{eq-arcface} is that the gradient $\nabla_{\theta} \mathcal{L}_A$ is proportional to the gradient $\nabla_{\theta} \tilde f_{\theta}$, where $\tilde f_{\theta}  \doteq f_{\theta}/\|f_{\theta}\|$ because of the $\ell_2$ normalization of the softmax inputs. Since $\tilde f_{\theta}$ lives on the unit hypersphere, the gradient $\nabla_{\theta} \tilde f_{\theta}$ will always be tangent to it. Therefore, no gradients perpendicular to the hypersphere will be back-propagated to supervise $f_{\theta}$ for reducing the intra-class variability of the embedding, while maximizing the inter-class discrepancy. This issue suggests that adding a regularizing term to the loss~\eqref{eq-arcface}, which allows orthogonal gradients to flow back could increase the hypothesis space exploration of the embedding  $f_{\theta}$, and better become invariant to nuisance factors of variation.

By adding~\eqref{eq-batch-hard} to~\eqref{eq-arcface} as in~\eqref{eq-joint-loss}, we are addressing the issue highlighted above. Indeed, the intent of~\eqref{eq-batch-hard} and~\eqref{eq-arcface} is the same, but in~\eqref{eq-batch-hard} we do not have the requirement for $f_{\theta}$ to be $\ell_2$ normalized. Hence, by picking a distance $D(\cdot,\cdot)$ that does not normalize the embedding, \eqref{eq-joint-loss} enables the gradient to flow in all directions. In~\cite{deng2018arcface} they attempted merging~\eqref{eq-arcface} with~\eqref{eq-triplet} without success, but there they used a distance with a normalized embedding, which we advocate not to use in this case. In our experiments, we picked $D(\cdot,\cdot)$ to be the Euclidean distance.

The second reason for using~\eqref{eq-joint-loss} comes from the composition of a batch, which has certain requirements because of the batch hard mining. We note that the same batch made of $N=PK$ samples can actually be used for the loss~\eqref{eq-arcface}. More importantly, this approach may prevent issues related to imbalanced data. Since datasets for person re-identification may have identities with a lot more samples than others, sampling a constant number of identities, from which we sample a constant number of images, imposes the embedding to be trained uniformly across the identities, rather than being under/over trained on some of them. In all of our experiments we sample the batches as it is done for the batch hard mining, regardless of the loss that we use.

Moreover, we note that since the loss~\eqref{eq-batch-hard} exercises a set of push-pull forces, it might be that when used as in~\eqref{eq-joint-loss}, the effect of the additive angular margin in~\eqref{eq-arcface} could become less relevant. Indeed, this is one of our conclusions.

In addition to~\eqref{eq-joint-loss}, we also replace the classification loss with the multi-task loss including the identity component $\mathcal{L}_{AM_{id}}  (\theta_{id}, W_{id})$, and the attribute component~\eqref{eq-arcface-attr}. This leads to the full re-identification training model, given by
\begin{equation}
  \mathcal{L}_{AMBH_{Attr}} (\theta,W) = \mathcal{L}_{AM}(\theta_{id},W_{id}) + \lambda \mathcal{L}_{AM_{Attr}}(\theta_a,W_a)
   + \gamma \mathcal{L}_{BH} (\theta_{id}) \; ,
  \label{eq-joint-loss_attribute}
\end{equation}
where $\lambda$ and $\gamma$ are hyperparameters that stryke a balance between the identity, the attributes, and the metric learning terms.

\subsection{Network Architecture}

As in most of the recent literature on person re-identification, we use a pretrained ResNet-50~\cite{He2016cvpr} as backbone network. We simply discard the fully connected layer, and we change the stride of the last convolutional stage from 2 to 1. We then add a global average pooling (GAP) layer and a batch normalization layer (BN). The dimensionality of the embedding features is 2048. At this point weight normalization and $\ell_2$ feature normalization is applied before entering the additive angular margin softmax loss~\eqref{eq-arcface-attr}, while no normalization is needed for the batch hard triplet loss component~\eqref{eq-batch-hard}. Figure~\ref{fig:architecture} is a simple exemplification of the architecture. During testing, unless otherwise specified, we perform all the experiments with the $\ell_2$ normalized embedding $f_{\theta}/\|f_{\theta}\|$, and re-identification is done via cosine similarity. Specifically $\theta$ is actually $\theta_{id}$, when the network has been trained with the full model~\eqref{eq-joint-loss_attribute}.


\section{Experiments}
\label{sec:experiments}
\begin{table}[t!]
  \centering
  	\caption{Comparison with the sate-of-the-art methods on Market-1501 and DukeMTMC-reID. The best and second best are shown in red and blue respectively.}
	\label{table-main-market-duke}
	\begin{tabular}{|l|c|c|c|c|c|}
		
		\multicolumn{2}{c} {} & \multicolumn{2}{c}{ \textbf{Market-1501}} & \multicolumn{2}{c}{ \textbf{DukeMTMC-reID}} \\
		\hline
		Method & Backbone & Rank-1 & mAP & Rank-1 & mAP  \\
		\hline\hline
		FD-GAN~\cite{ge2018fd} & ResNet & 90.5 & 77.7 & 80.0 & 65.4\\
		Part-aligned~\cite{suh2018part} & GoogleNet & 91.7 & 79.6 & 84.4 & 69.3\\
		SGGNN~\cite{shen2018person} & ResNet & 92.3 & 82.8 & 81.1 & 68.2\\
		PCN+PCP~\cite{chikontwe2018deep} & ResNet & 92.8 & 78.8 & 85.7 & 71.2\\
		Mancs~\cite{wang2018mancs} & ResNet & 93.1 & 82.3 & 84.9 & 71.8\\
		APDR~\cite{li2020attributes} & ResNet & 93.1 & 80.1 & 84.3 & 69.7\\
		DeepCRF~\cite{chen2018group} & ResNet & 93.5 & 81.6 & 84.9 & 69.5\\
		PCB~\cite{sun2018beyond} & ResNet & 93.8 & 81.6 & 83.3 & 69.2\\
		AA-Net~\cite{tay2019aanet} & ResNet & 93.9 & 83.4 & 87.7 & 74.3\\
		IA-Net~\cite{hou2019interaction} & ResNet & 94.4 & 83.1 & 87.1 & 73.4\\
		SphereReID~\cite{fan2019spherereid} & ResNet & 94.4 & 83.6 & 83.9 & 68.5\\
		CAMA~\cite{yang2019towards} & ResNet & {94.7} & 84.5 & 85.8 & 72.9\\
		DG-Net~\cite{zheng2019joint} & ResNet & \s{94.8} & \s{86.0} & 86.6 & 74.8\\
		\hline
\textbf{AM0BH} (Ours) & ResNet & 94.6\std{0.21} & 85.9 \std{0.28} & \s{89.2}\std{0.40} & \s{76.7}\std{0.26}\\
\textbf{AM0BH$_{\textbf{Attr}}$} (Ours) & ResNet & \f{94.9}\std{0.13} & \f{86.3}\std{0.10} & \f{89.3}\std{0.19} & \f{77.4}\std{0.14}\\
		\hline
	\end{tabular}
      \end{table}
      
We evaluate our model on three person re-identification datasets. Every evaluation was repeated 10 times. We report the average performance metrics with their standard deviations for the following datasets.

\textbf{Market-1501:} contains 32668 images of 1501 identities captured by six cameras ~\cite{zheng2015scalable}.

\textbf{DukeMTMC-reID:} contains 36441 images of 1812 identities captured by eight high resolution cameras ~\cite{zheng2017unlabeled}.

\textbf{MSMT17:} is the most recent and challenging person re-identification dataset. It contains 126441 images of 4101 identities captured by 15 cameras~\cite{wei2018person}.

We use the data provided in~\cite{lin2019improving} as attribute labels for Market-1501 and DukeMTMC. These attributes are manually annotated at the identity level. There are 27 attributes for  Market-1501 and 23 attributes for DukeMTMC. Some examples of attributes include: gender, hair length, carrying backpack, carrying handbag, wearing hat, different upper body and lower body clothing colors, length and type of lower body clothing, shoe type and shoes color.

\subsection{Implementation Details}
\label{subsec:implementation-details}

We implemented our approach with PyTorch~ \cite{paszke2017automatic}, and for the backbone network we use ResNet-50~\cite{he2016deep} pre-trained on ImageNet~\cite{deng2009imagenet}. The batch size is 32 where $P=4$ and $K=8$. For the Market-1501 dataset the input image size is $256 \times 128$ while it is $288 \times 144$ for DukeMTMC-reID and the MSMT17 datasets.

For data augmentation, training images are randomly flipped and erased. The model is trained using the Adam optimizer with default hyper parameters for 150 epochs. The learning rate is linearly increased from $10^{-5}$ to $10^{-3}$ for the first 20 epochs to help the network bootstrap. Then the learning rate is set to $10^{-3}$ after the first 20 epochs, and it decreases to $10^{-4}$ and $10^{-5}$ after epochs 90, and 130, respectively.

There are two settings for our proposed approach. The first one, AM0BH, uses the joint loss~\eqref{eq-joint-loss} with no attributes. 
$\gamma$ is 0.43, 0.5 and 0.4 for Market-1501, DukeMTMC-reID and MSMT17, respectively. The second setting, AM0BH$_{Attr}$, leverages the full model~\eqref{eq-joint-loss_attribute}.
The 2048 embedding features are divided into $f_{\theta_{a}}$ and $f_{\theta_{id}}$. $f_{\theta_{a}}$ has size $ M \times Q$, where $M$ is the total number of attributes and $Q$ is the size of the input features to each of the attribute classifiers. $Q$ is set to 16. The rest of the embedding features, $f_{\theta_{id}}$, is the input to the identity classifier and triplet loss. 
In~\eqref{eq-joint-loss_attribute} we set $\lambda$ to $0.25$ and $\gamma$ to $0.54$ for Market-1501. While for DukeMTMC-reID, we set $\lambda$ to $0.2$ and $\gamma$ to $0.33$.
\begin{table}[t!]
	\centering
	\caption{Comparison with the-sate-of-the-art methods on the MSMT17 dataset. The best and second best are shown in red and blue respectively.}
	\label{table-main-msmt}
	\begin{tabular}{|l|c|c|c|c|c|}
		\hline
		Method & Backbone & Rank-1 & Rank-5 & Rank-10 & mAP \\
		\hline\hline
		PCB~\cite{sun2018beyond} & ResNet & 68.2 & 81.2 & 85.5 & 40.4\\
		IA-Net~\cite{hou2019interaction} & ResNet & 75.5 & 85.5 & 88.7 & 46.8\\
		DG-Net~\cite{zheng2019joint} & ResNet & \s{77.2} & \s{87.4} & \s{90.5} & \s{52.3}\\
		\hline
\textbf{AM0BH} (Ours)   & ResNet & \f{78.1}\std{0.40}  &\f{88.3}\std{0.13} & \f{91.2}\std{0.19} & \f{53.4}\std{0.32}\\
		\hline
	\end{tabular}
\end{table}

\subsection{Comparison with State-of-the-Art Methods}
\label{subsec:comparison-with-state-of-the-art-methods}

The performance is evaluated by CMC (Cumulative Matching Characteristic) and mAP (Mean Average Precision) after computing the matching score between the probe image and gallery images. We discard the score if the probe image and galley image are from the same view. 

To show the performance of our proposed approach, we compare it with the state-of-the-art methods on three person re-identification dataset. However, we are not able to implement AM0BH$_{Attr}$ on MSMT17 since there is no attributes annotation available.
Table~\ref{table-main-market-duke} shows the results on Market-1501 and DukeMTMC-reID. For Market-1501, AM0BH outperforms most of the state-of-the-art and the performance is close to the best, i.e., DG-Net~\cite{zheng2019joint} in terms of rank-1 and mAP. While for DukeMTMC-reID, AM0BH outperforms the state-of-the-art by achieving 89.2\% on rank-1 and 76.7\% mAP.
Table~\ref{table-main-msmt} shows results on MSMT17. The proposed AM0BH outperforms DG-Net~\cite{zheng2019joint} by a gap of 0.9\%, 0.9\%, 0.7\% and 1.1\% for rank-1, rank-5, rank-10 and mAP respectively. By using attributes, AM0BH$_{Attr}$ further improves the performance over AM0BH by 0.3\% and 0.4\% for rank-1 and mAP for Market-1501, and by 0.1\% and 0.7\% for rank-1 and mAP for DukeMTMC-reID.

\renewcommand{\std}[1]{\tiny {$\pm$#1} }  

\begin{table}[t!]
  \centering
  	\caption{\textbf{Ablation study}.
		Shows the effect of different loss combinations.
		Losses: a) AM0 - softmax loss~\eqref{eq-arcface} when margin is set to $0$; b) AM - softmax loss~\eqref{eq-arcface} when margin is set to $0.5$; c) BH - batch hard triplet loss~\eqref{eq-batch-hard};
		d) AM0BH1 - softmax loss when margin is set to $0$ combined with batch hard triplet loss with feature normalization;
		e) AMBH - softmax loss when margin is set to $0.5$ combined with batch hard triplet loss;
		f) AM0BH - softmax loss when margin is set to $0.0$ combined with batch hard triplet loss;
		g) AM0BHsp - softmax loss when margin is set to $0.0$ combined with batch hard triplet loss with softplus function instead of hinge loss.
	}
	\label{table-ablation-losses}
        \begin{adjustbox}{width=\columnwidth}
	\begin{tabular}{lcccc|cccc|cccc}
        \toprule
		\small{\textbf{Loss}} & \multicolumn{4}{c}{ \small \textbf{Market-1501}} & \multicolumn{4}{c}{ \small \textbf{DukeMTMC-reID}}& \multicolumn{4}{c}{ \small \textbf{MSMT17}}  \\
		\cmidrule(l{2pt}r{2pt}){2-5} \cmidrule(l{2pt}r{2pt}){6-9} \cmidrule(l{2pt}r{2pt}){10-13}
		  & Rank1 & Rank5 & Rank10  & mAP & Rank1 & Rank5 & Rank10  & mAP & Rank1 & Rank5 & Rank10  & mAP \\
        AM0              & 92.86\std{0.34}    & 97.57\std{0.12}    & 98.47\std{0.04}    & 83.67\std{0.17}    & 87.74\std{0.29}     & 94.06\std{0.17}     & 95.62\std{0.35}    & 74.60\std{0.30}   & 77.50\std{0.25}     & 87.80\std{0.30}     & 90.78\std{0.18}    & 51.82\std{0.19} \\
        AM               & 94.16\std{0.23}    & 98.04\std{0.21}    &\s{98.92}\std{0.08} & 84.54\std{0.22}    & 88.31\std{0.18}     & 94.34\std{0.48}    & 95.78\std{0.24}    & 75.51\std{0.16}    &\s{78.20}\std{0.28}  & 88.15\std{0.42}	 & 91.15\std{0.35}    & 53.00\std{0.49} \\
        BH               & 84.74\std{0.23}    & 94.64\std{0.36}    & 96.74\std{0.13}    & 67.40\std{0.46}    & 81.60\std{0.49}     & 91.02\std{0.38}    & 93.46\std{0.18}    & 65.16\std{0.31}    & 56.34\std{0.92}     & 73.26\std{0.99}    & 79.30\std{0.70}    & 30.86\std{0.67}\\
        AM0BH1           & 94.28\std{0.13}    &\s{97.90}\std{0.20} & 98.80\std{0.07}    & 84.52\std{0.16}    & 88.02\std{0.18}     & 93.96\std{0.27}    & 95.48\std{0.26}    & 74.56\std{0.32}    & 77.46\std{0.13}     & 87.64\std{0.15}    & 90.60\std{0.19}    & 51.86\std{0.22}\\

        AMBH             & 93.29\std{0.40}    & 97.80\std{0.11}    & 98.70\std{0.14}    & 84.00\std{0.09}    & 88.18\std{0.46}     & 94.62\std{0.17}    & 96.19\std{0.18}    &\s{76.71}\std{0.21} & 77.93\std{0.47}     & 88.07\std{0.47}    & 91.07\std{0.38}    & 52.70\std{0.53} \\
		\midrule
\textbf{AM0BH}(Ours)      &\f{94.64}\std{0.21} &\f{98.22}\std{0.16} &\f{99.02}\std{0.11} &\f{85.90}\std{0.28} &\f{89.20}\std{0.40}  &\s{94.72}\std{0.33} &\s{96.26}\std{0.17} & 76.68\std{0.26}    &78.14\std{0.40}      &\s{88.34}\std{0.13}    &\f{91.24}\std{0.19} &\f{53.44}\std{0.32}\\
\textbf{AM0BHsp}(Ours)    &\s{94.42}\std{0.15} &\f{98.22}\std{0.19} &\f{99.02}\std{0.13} &\s{85.76}\std{0.19} &\s{88.79}\std{0.31}  &\f{94.85}\std{0.22} &\f{96.32}\std{0.19} &\f{77.42}\std{0.3}  &\f{78.26}\std{0.27}  &\f{88.38}\std{0.11} & \s{91.20}\std{0.14}    &\s{53.36}\std{0.40} \\
        \bottomrule
	\end{tabular}
        \end{adjustbox}
\end{table}

\subsection{Ablation Study}
\label{subsec:ablation-study}

In the ablation study we compare different loss combinations on all three datasets used in Section~\ref{subsec:comparison-with-state-of-the-art-methods}. First, we examine how the identity classification loss~\eqref{eq-arcface}, and the metric loss~\eqref{eq-batch-hard} behave independently. The summary results of this experiment are included in Table~\ref{table-ablation-losses}. Then we examine different combinations.

\textbf{Identity classification loss}.
We start by examining the additive angular loss applied for identity classification~\eqref{eq-arcface}. When the margin is set to $0$ (row AM0 in Table~\ref{table-ablation-losses}), this case is equivalent to the loss described in~\cite{fan2019spherereid}.
Then, we use the loss~\eqref{eq-arcface} when the margin is set to $0.5$, as in~\cite{deng2018arcface} (row AM in the Table~\ref{table-ablation-losses}).
We observe a significant improvement of all metrics, which proves that the additive angular margin has a positive effect when the softmax loss is used alone.

\textbf{Metric learning loss}.
We continue by examining batch hard triplet loss~\eqref{eq-batch-hard}. This is the row BH in Table~\ref{table-ablation-losses}. It can be seen that this loss alone underperforms the identity classification losses in rows AM, and AM0.
\begin{figure}[t!]
\begin{center}
	
	\includegraphics[width=0.8\linewidth]{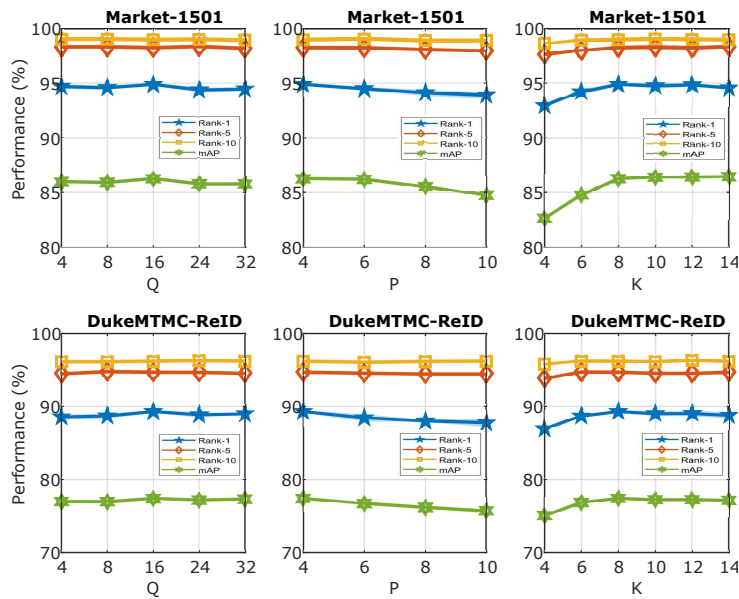}

\end{center}
\caption{\textbf{Ablation study}.
		Shows the effect of $Q$, $P$ and $K$ on the performance of AM0BH$_{\textbf{Attr}}$. $Q$ is the size of embedding features fed to each attribute classifier, $P$ is the number of identities in each mini-batch ($P$ is fixed to 4) and $K$ is the number of samples per identity in each mini-batch ($K$ is fixed to 8).}
\label{fig:ablation-QPK-performance}
\end{figure}

\textbf{Identity classification and metric learning losses}.
We analyze four cases: the combination of AM0 and BH, the combination of AM and BH, the combination of AM0 and BH with feature normalization for BH, and the combination of AM0 and BH with softplus instead of hinge loss.

\emph{Combination of AM0 and BH}.
The combination of the softmax loss~\eqref{eq-arcface} with margin set to 0 and batch hard triplet loss~\eqref{eq-batch-hard} is presented in the row AM0BH of Table~\ref{table-ablation-losses}.
We observe improvement on almost all metrics compared to AM0 or BH individually, which signifies that they complement each other. 

\emph{Combination of AM and BH}.
The combination of the additive angular margin softmax loss~\eqref{eq-arcface} with margin set to $0.5$ and the batch hard triplet loss~\eqref{eq-batch-hard} is presented in the row AMBH of Table~\ref{table-ablation-losses}.
We observe that it does not provide improvement compared to AM0BH. This means that the angular margin becomes less relevant, since the batch hard triplet loss exercises a set of push-pull forces that is likely comparable to the effect of the margin, and might even generate conflicts when this is present, leading in this case, to results closer to AM alone.

\emph{Combination of AM0 and BH with normalized features}.
The combination of the softmax loss~\eqref{eq-arcface} with margin set to $0$, and the batch hard triplet loss~\eqref{eq-batch-hard} with normalized features is presented in row AM0BH1 of Table~\ref{table-ablation-losses}. We note a decreased performance, when compared with row AM0BH. As described in Section~\ref{subsec:joint-class-metric}, this might be due to the feature normalization prior to the triplet loss, which forces no gradients perpendicular to the hypersphere to be back-propagated to supervise $f_{\theta}$. Removing that constraint would allow the orthogonal gradients flow that could increase the hypothesis space exploration of the embedding $f_{\theta}$. 

\emph{Combination of AM0 and BH with softplus}.
The combination of the softmax loss~\eqref{eq-arcface} with margin set to $0$ and batch hard triplet loss~\eqref{eq-batch-hard} with softplus function instead of the hinge loss is presented in row AM0BHsp of Table~\ref{table-ablation-losses}.
It shows overall similar performance to AM0BH, but slightly higher mAP, while slightly lower rank1-10 metrics.
We speculate that hinge loss concentrates only on the triplets within the margin, ignoring the tail of the distribution, which is beneficial for rank-1 - rank-10 metrics, while with softplus the whole distribution of triples is accounted in the loss, which is beneficial for the mAP metric.

\textbf{Identity and attribute classification with metric learning losses}.
Here we present the ablation study that supports the addition of the attribute classification task to further improve the robustness against nuisance factors, as previously suggested. We study the influence of $Q$ and batch size on the performance of AM0BH$_{Attr}$. Figure~\ref{fig:ablation-QPK-performance} shows the effect of the size, $Q$, of the embedding features fed to each attribute classifier. The best performance is achieved when $Q$ is 16. Figure~\ref{fig:ablation-QPK-performance} also shows the effect of the batch size by examining different values of $P$ and $K$ respectively on the performance.


\section{Conclusions}

We have further studied the learning of a feature embedding for person re-identification via a joint optimization of a discriminative and a metric learning loss to minimize the intra-class variation and maximize the inter-class separation. Our approach was motivated by observing untapped limitations imposed by a margin based softmax loss onto the gradient flow that supervises the training of the embedding. We have verified that adding a triplet loss as regularizer serves as proxy for the missing gradient directions, and enables learning a better embedding. Moreover, we have shown that adding a discriminative semantic task like predicting attributes, further strengthens the robustness of the representation. We have verified that on the three most challenging datasets by setting new state-of-the-art performance for the case of holistic representations for person re-identification that do not leverage explicit modeling of nuisance factors (e.g., pose). Moreover, we found that the joint loss achieves its best performance when we do not require a margin in the softmax portion, showing the importance of the contribution added by the triplet component, when it is used to expand the directions of the gradient flow.

\subsubsection{Acknowledgements} This material is based upon work supported in part by the National Science Foundation under Grant No. 1920920.

%
%
%
\bibliographystyle{splncs04}
\bibliography{references}

\end{document}